\documentclass[a4paper,12pt]{article}
\usepackage{amsmath}
\usepackage{amssymb}
\usepackage{amsfonts}
\usepackage{latexsym}
\usepackage{enumerate}
\usepackage{stmaryrd}

\newtheorem{llemma}{Lemma}[section]
\newtheorem{prop}[llemma]{Proposition}

\newtheorem{exmp}[llemma]{Example}
\newtheorem{thm}[llemma]{Theorem}

\newtheorem{defn}[llemma]{Definition}

\newtheorem{key}[llemma]{Keyword}

\newcommand{\proof}{\emph{Proof. }}

\begin{document}
\title{\textbf{Neutrosophic soft sets with applications in decision making}}
\author{Faruk Karaaslan}
\date{\footnotesize{Department of Mathematics, Faculty of Science,
 \c{C}ank{\i}r{\i} Karatekin University, 18100 \c{C}ank{\i}r{\i},
 Turkey\\
           fkaraaslan@karatekin.edu.tr}}

\maketitle

%***********************************************************************************
% ABSTRACT
\begin{abstract} We firstly present definitions and  properties in study of Maji \cite{maji-2013}
on neutrosophic soft sets. We then give a few notes on his study.
Next, based on \c{C}a\u{g}man \cite{cagman-2014}, we redefine the
notion of neutrosophic soft set and neutrosophic soft set operations
to make more functional. By using these new definitions we construct
a decision making method  and a group decision making method which
selects a set of optimum elements from the alternatives. We finally
present  examples which shows that the methods can be successfully
applied to many problems that contain uncertainties.
%-----------------------------------------------------------------
\begin{key} Neutrosophic set; Soft set;
Neutrosophic soft set; decision making.
\end{key}
\end{abstract}
%***********************************************************************************
%% section 1: introduction

\section{Introduction}
%***********************************************************************************
Many problems including uncertainties are a major issue in many
fields of real life such as economics, engineering, environment,
social sciences, medical sciences and business management. Uncertain
data in these fields could be caused by complexities and
difficulties in classical mathematical modeling. To avoid
difficulties in dealing with uncertainties, many tools have been
studied by researchers. Some of these tools are fuzzy sets
\cite{zadeh-1965}, rough sets \cite{paw-82} and intuitionistic fuzzy
sets \cite{atanassov-1986}. Fuzzy sets and  intuitionistic fuzzy
sets  are characterized by membership functions, membership and
non-membership functions, respectively.  In some real life problems
for proper  description of  an object in uncertain and ambiguous
environment, we need to handle the indeterminate and incomplete
information. But fuzzy sets and  intuitionistic fuzzy sets don't
handle the indeterminant and inconsistent  information. Samarandache
\cite{smar-95} defined the notion of neutrosophic set  which is a
mathematical tool for dealing with problems involving imprecise and
indeterminant data.

Molodtsov introduced concept of soft sets \cite{molodtsov-1999} to
solve complicated problems and various types of uncertainties. In
\cite{maj-03sst}, Maji et al. introduced several operators for soft
set theory: equality of two soft sets, subsets and superset of  soft
sets, complement of soft set, null soft sets and absolute soft sets.
But some of these definitions and their properties have   few gaps,
which have been pointed  out by Ali et al.\cite{ali-09osnop} and
Yang \cite{yang-08}. In 2010, \c{C}a\u{g}man and Engino\u{g}lu
\cite{cag-10sstuidcm} made some modifications  the operations of
soft sets and filled in these gap. In 2014, \c{C}a\u{g}man
\cite{cagman-2014} redefined soft sets using the single parameter
set and compared definitions with those defined before.

Maji \cite{maji-2013} combined the concept of  soft set and
neutrosophic set together by introducing a new concept called
neutrosophic soft set and gave  an application of neutrosophic soft
set in decision making problem. Recently, the properties and
applications on the neutrosophic  sets have been studied
increasingly \cite{bro-13gns,bro-13ins,deli-14ivnss,deli-14nsm}.The
propose of this paper is to fill the gaps of the Maji's neutrosophic
soft set \cite{maji-2013}  definition and operations redefining
concept of neutrosophic soft set and operations between neutrosophic
soft sets. First, we present Maji's definitions and operations and
we verify that some propositions are incorrect by a counterexample.
Then based on \c{C}a\u{g}man's \cite{cagman-2014} study we redefine
neutrosophic soft sets and their operations. Also, we investigate
properties of  neutrosophic soft sets operations. Finally we present
an application of a neutrosophic soft set in decision making.

%% Section 2: preliminaries
\section{Preliminaries}
In this section, we will recall the notions of neutrosophic sets
\cite{smarandache-2005} and soft sets \cite{molodtsov-1999}. Then,
we will give some properties of these notions. Throughout this paper
$X$, $E$ and $P(X)$  denote initial universe, set of parameters and
power set of $X$, respectively.
%-----------------------------------------------------------------------------------
\begin{defn}\cite{smarandache-2005}
A neutrosophic set $A$ on the universe of discourse $X$ is defined as
$$
A=\big\{\langle x,T_A(x),I_A(x),F_A(x)\rangle: x\in X\big\}
$$
where $T_A,I_A,F_A: X \to]^-0,1^+[$ and $^-0\leq
T_A(x)+I_A(x)+F_A(x)\leq3^+$. From philosophical point of
view, the neutrosophic set takes the value from real standard or
non-standard subsets of $]^-0,1^+[$. But in real life application in
scientific and engineering problems it is difficult to use
neutrosophic set with value from real standard or non-standard
subset of $]^-0,1^+[$. Hence we consider the neutrosophic set which
takes the value from the subset of $[0,1]$.
\end{defn}
%-----------------------------------------------------------------------------------
\begin{defn}\cite{molodtsov-1999}
Let consider a nonempty set $A$, $A\subseteq E$. A pair $(F,A)$ is called a soft set
over $X$, where $F$ is a mapping given by $F:A\to P(X)$.
\end{defn}
%-----------------------------------------------------------------------------------
\begin{exmp}\label{e-softset}
Let $X=\{x_l,x_2,x_3,x_4,x_5,x_6,x_7,x_8\}$ be the universe which are eight houses
and $E=\{e_1,e_2,e_3,e_4,e_5,e_6\}$ be the set of parameters. Here, $e_i$ $(i=1,2,3,4,5,6)$
stand for the parameters ``\emph{modern}'', ``\emph{with parking}'', ``\emph{expensive}'',
``\emph{cheap}'', ``\emph{large}'' and ``\emph{near to city}'' respectively. Then,
following soft sets are described respectively Mr. A and Mr. B  who are going to buy
\begin{eqnarray*}
F & = & \big\{(e_1,\{x_1,x_3,x_4\}),(e_2,\{x_1,x_4,x_7,x_8\}),(e_3,\{x_1,x_2,x_3,x_8\})\big\}\\
G & = & \big\{(e_2\{x_1,x_3,x_6\}),(e_3,X),(e_5,\{x_2,x_4,x_4,x_6\})\big\}.
\end{eqnarray*}
\end{exmp}
%-----------------------------------------------------------------------------------
From now on, we will use definitions and  operations of soft sets
which are more suitable for pure mathematics  based on study of  \c{C}a\u{g}man \cite{cagman-2014}.
%-----------------------------------------------------------------------------------
\begin{defn}\cite{cagman-2014}
A soft set $F$ over $X$ is a set valued function from $E$ to $P(X)$.  It can
be written a set of ordered pairs
$$
F=\big\{(e,F(e)):e\in E\big\}.
$$
Note that if $F(e)=\emptyset$, then the element $(e, F(e))$ is not appeared in $F$.
Set of all soft sets over $X$ is denoted by $\mathbb{S}$.
\end{defn}
%-----------------------------------------------------------------------------------
\begin{defn}\cite{cagman-2014} Let $F,G\in\mathbb{S}$. Then,
\begin{enumerate}[\it i.]
\item If $F(e)=\emptyset$ for all $e\in E$, $F$ is said to be a null soft set, denoted
by $\Phi$.
\item If $F(e)=X$ for all $e\in E$, $F$ is said to be absolute soft set, denoted by
$\hat{X}$.
\item $F$ is soft subset of $G$, denoted by $F\tilde\subseteq G$, if $F(e)\subseteq G(e)$
for all $e\in E$.
\item $F=G$, if $F\tilde\subseteq G$ and $G\tilde\subseteq F$.
\item Soft union of $F$ and $G$, denoted by $F\tilde\cup G$, is a soft set over $X$ and
defined by $F\tilde\cup G:E\to P(X)$ such that $(F\tilde\cup G)(e)=F(e)\cup G(e)$
for all $e\in E$.
\item Soft intersection of $F$ and $G$, denoted by $F\tilde\cap G$, is a soft set over
$X$ and defined by $F\tilde\cap G:E\to P(X)$ such that $(F\tilde\cap G)(e)=F(e)\cap G(e)$
for all $e\in E$.
\item Soft complement of $F$ is denoted by $F^{\tilde c}$ and defined by
$F^{\tilde c}:E\to P(X)$ such that $F^{\tilde c}(e)=X\setminus F(e)$ for all $e\in E$.
\end{enumerate}
\end{defn}
%-----------------------------------------------------------------------------------
\begin{exmp}
Let us consider soft sets $F$,$G$ in the Example \ref{e-softset}. Then, we have
\begin{eqnarray*}
F\tilde\cup G & = & \big\{(e_1,\{x_1,x_3,x_4\}),(e_2,\{x_1,x_3,x_4,x_6,x_7,x_8\}),\\
              &   & (e_3,X),(e_5,\{x_2,x_4,x_4,x_6\})\big\}\\
F\tilde\cap G & = & \big\{(e_2\{x_1\}),(e_3,\{x_1,x_2,x_3,x_8\})\big\}   \\
F^{\tilde c}  & = & \big\{(e_1,\{x_2,x_5,x_6,x_7,x_8\}),(e_2,\{x_2,x_3,x_5,x_6\}),\\
& & (e_3,\{x_4,x_5,x_6,x_7\}),(e_4,X),(e_5,X),(e_6,X)\big\}.
\end{eqnarray*}
\end{exmp}
%------------------------------------------------------------------------
\begin{defn} \cite{maji-2013} Let $X$ be an initial universe set and
$E$ be a set of parameters. Consider $A\subset E$. Let $P(X)$
denotes the set of  all neutrosophic sets of $X$. The collection
$(F,A)$ is termed to be the soft neutrosophic set over $X$, where
$F$ is a mapping given by $F:A \to P(X).$
\end{defn}
For illustration we consider an example.
%------------------------------------------------------------------------
\begin{exmp}\label{examplemaji}
Let $X$ be the set of houses under consideration and $E$ is the set of
parameters. Each parameter is a neutrosophic word or sentence
involving neutrosophic words. Consider $E=\{ \textrm{beautiful,
wooden, costly, very costly},$  $\textrm{moderate, green
surroundings, in good repair, in bad repair, cheap, expensive}\}$.
In this case, to define a neutrosophic soft set means to point out
beautiful houses, wooden houses, houses in the green surroundings
and so on. Suppose that, there are five houses in the universe $X$
given by, $U = \{h_1, h_2, h_3, h_4, h_5\}$ and the set of
parameters $A = \{e_1, e_2, e_3, e_4\}$, where $e_1$ stands for the
parameter 'beautiful', $e_2$ stands for the parameter 'wooden',
$e_3$ stands for the parameter 'costly' and the parameter $e_4$
stands for 'moderate'. Suppose that,
\begin{eqnarray}
F(beautiful) &=& \{\langle h_1, 0.5, 0.6, 0.3 \rangle,\langle  h_2,
0.4, 0.7, 0.6 \rangle,\langle  h_3, 0.6, 0.2, 0.3 \rangle,\nonumber\\
&& \langle h_4, 0.7, 0.3, 0.2 \rangle,\langle h_5, 0.8, 0.2, 0.3
\rangle\},\nonumber\\
F(wooden) &=& \{\langle h_1, 0.6, 0.3, 0.5 \rangle,\langle h_2, 0.7,
0.4, 0.3 \rangle, \langle h_3, 0.8, 0.1, 0.2 \rangle,\nonumber\\
&&\langle h_4, 0.7, 0.1, 0.3 \rangle,\langle h_5, 0.8, 0.3, 0.6
\rangle\},\nonumber\\
F(costly) &=& \{\langle h_1, 0.7, 0.4, 0.3 \rangle,\langle h_2, 0.6,
0.7, 0.2 \rangle, \langle h_3, 0.7, 0.2, 0.5 \rangle,\nonumber \\
&&\langle h_4, 0.5, 0.2, 0.6 \rangle, \langle h_5, 0.7, 0.3, 0.4
\rangle \},\nonumber\\
F(moderate) &=& \{\langle h_1, 0.8, 0.6, 0.4 \rangle,\langle h_2,
0.7, 0.9, 0.6 \rangle,\langle h_3, 0.7, 0.6, 0.4 \rangle,
\nonumber\\
&&\langle h_4, 0.7, 0.8, 0.6 \rangle,\langle h_5, 0.9, 0.5, 0.7
\rangle\}.\nonumber
\end{eqnarray}
The neutrosophic soft set $(NSS)$ $(F,E)$ is a parameterized family
$\{F(e_i); i = 1,2,...,10\} $of all neutrosophic sets of $X$ and
describes a collection of approximation of an object.

Thus we can view the neutrosophic soft set $(NSS)$ $(F,A)$ as a
collection of approximation as below:
\begin{eqnarray*}
(F,A)&=&\{ beautiful\; houses = \{\langle h_1, 0.5, 0.6, 0.3
\rangle,\langle h_2, 0.4, 0.7, 0.6 \rangle,\\
&& \langle h_3, 0.6, 0.2,
0.3 \rangle,\langle h_4, 0.7, 0.3, 0.2 \rangle,\langle h5, 0.8, 0.2,
0.3 \rangle\},\\
&& wooden\; houses = \{\langle h_1, 0.6, 0.3, 0.5 \rangle,\langle h_2, 0.7, 0.4, 0.3
\rangle,\\
&&\langle h_3, 0.8, 0.1, 0.2 \rangle, \langle h_4, 0.7, 0.1,
0.3 \rangle,\langle h_5, 0.8, 0.3, 0.6 \rangle\}, \\
&&costly \;houses = f\langle h_1,
0.7, 0.4, 0.3 \rangle ,\langle h_2, 0.6, 0.7, 0.2 \rangle, \\
&&\langle h_3, 0.7, 0.2, 0.5 \rangle,\langle h_4, 0.5, 0.2, 0.6
\rangle,\langle h_5, 0.7, 0.3, 0.4 \rangle\},\\
&& moderate \;houses=\langle h_1, 0.8, 0.6, 0.4 \rangle,\langle h_2, 0.7, 0.9, 0.6
\rangle,\\
&&\langle h_3, 0.7, 0.6, 0.4 \rangle, \langle h_4, 0.7, 0.8,
0.6 \rangle,
\langle h_5, 0.9, 0.5, 0.7 \rangle\}\}.
\end{eqnarray*}
\end{exmp}
%------------------------------------------------------------------------
\begin{defn}\cite{maji-2013}\label{ns-subset} Let $(F,A)$ and $(G,B)$ be two
neutrosophic sets over the common universe $X$. $(F,A)$ is said to
be neutrosophic soft subset of $(G,B)$ is $A\subset B$, and
$T_{F(e)}(x)\leq T_{G(e)}(x)$, $ I_{F(e)}(x)\leq I_{G(e)}(x)$
$F_{F(e)}(x) \geq F_{G(e)}(x)$, $\forall e\in A$, $\forall x\in U$.
We denote it by $(F,A) \subseteq (G,B)$. $(F,A)$ is said to be
neutrosophic soft super set of $(G,B)$ if $(G,B)$ is a neutrosophic
soft subset of $(F,A)$. We denote it by $(F,A)\supseteq(G,B)$.

if $(F,A)$ is neutrosophic
soft subset of $(G,B)$ and $(G,B)$ is neutrosophic soft subset of
$(F,A)$. We denote it $(F,A)=(G,B).$
\end{defn}
%------------------------------------------------------------------------
\begin{defn}\cite{maji-2013}\label{ns-notset} NOT set of a parameters.
Let $E=\{e_1,e_2,...e_n\}$ be a set of parameters. The NOT set of
$E$, denoted by $\rceil E$ is defined by $\rceil E=\{\neg e_1, \neg
e_2,...\neg e_n\}$, where $\neg e_i=$not $e_i$  $\forall i$(it may
be noted that $\rceil$ and $\neg$ are different operators).
\end{defn}
%------------------------------------------------------------------------
\begin{defn}\cite{maji-2013}\label{ns-complement} Complement of a neutrosophic soft set
$(F,A)$ denoted by $(F,A)^c$ and   is defined as $(F,A)^c=(F^c,\rceil
A)$, where $F^c. \, \rceil A\to P(X)$ is mapping given by $F^c(\alpha)=$
neutrosophic soft complement with $T_{F^c(x)}=F_{F(x)}$,
$I_{F^c(x)}=I_{F(x)}$ and $F_{F^c(x)}=T_{F(x)}$.
\end{defn}
%------------------------------------------------------------------------
\begin{defn}\cite{maji-2013}\label{ns-empty} Empty or null neutrosophic soft
set with respect to a parameter. A neutrosophic soft set $(H,A)$
over the universe $X$ is termed to be empty or null neutrosophic
soft set with respect to the parameter $e$ if $T_{H(e)}(m)=0,
F_{H(e)}=0$ and  $I_{H(e)}(m)=0$ $\forall m\in X$, $\forall e\in A$

In this case the null neutrosophic  soft set $(NNSS)$ is denoted by
$\Phi_A$
\end{defn}
%-----------------------------------------------------------------------------------
\begin{defn}\cite{maji-2013}\label{ns-union} Union of two
neutrosophic soft sets. Let $(H,A)$ and $(G,B)$ be two $NSSs$ over
the common universe $X$. Then the union of $(H,A)$ and $(G,B)$ is
defined by $(H,A)\cup (G,B)=(K,C)$, where $C=A\cup B$ and the
truth-membership, indeterminacy-membership and falsity-membership of
$(K,C)$ are as follow.
\begin{eqnarray}
T_{K(e)}(m)&=&T_{H(e)}(m), if e\in A-B\nonumber\\
&=&T_{G(e)}(m), if e\in B-A\nonumber\\
&=&max(T_{H(e)}(m),T_{G(e)}(m)), if e\in A\cap B\nonumber\\
I_{K(e)}(m)&=&I_{H(e)}(m), if e\in A-B\nonumber\\
&=&I_{G(e)}(m), if e\in B-A\nonumber\\
&=&\frac{I_{H(e)}(m)+I_{G(e)}(m)}{2} , if e\in A\cap B.\nonumber\\
F_{K(e)}(m)&=&F_{H(e)}(m), if e\in A-B\nonumber\\
&=&F_{G(e)}(m), if e\in B-A\nonumber\\
&=&min(F_{H(e)}(m),F_{G(e)}(m)), if e\in A\cap B\nonumber
\end{eqnarray}
\end{defn}
%----------------------------------------------------------------------------------
\begin{defn}\cite{maji-2013}\label{ns-intersection}  Let $(H,A)$ and $(G,B)$ be two $NSSs$ over
the common universe $X$. Then,  intersection of $(H,A)$ and
$(G,B)$ is defined by $(H,A)\cap (G,B)=(K,C)$, where $C=A\cap B$ and
the truth-membership, indeterminacy-membership and
falsity-membership of $(K,C)$ are as follow.
\begin{eqnarray}
T_{K(e)}(m)&=&min(T_{H(e)}(m),T_{G(e)}(m)), \; if\; e\in A\cap B\nonumber\\
I_{K(e)}(m)&=&\frac{I_{H(e)}(m)+I_{G(e)}(m)}{2},\; if\; e\in A\cap B.\nonumber\\
F_{K(e)}(m)&=&max(F_{H(e)}(m),F_{G(e)}(m)),\; if \;e\in A\cap B\nonumber
\end{eqnarray}
\end{defn}
For any two $NSSs$ $(H,A)$ and $(G,B)$ over the same universe $X$
and on the basis of the operations defined above, we have the
following propositions.
%----------------------------------------------------------------------------------
\begin{prop}\cite{maji-2013}\label{prop1}
\begin{enumerate}[(1)]
    \item $(H,A)\cup (H,A)=(H,A)$
    \item $(H,A)\cup (G,B)=(G,B)\cup (H,A) $
    \item $(H,A)\cap (H,A)=(H,A)$
    \item $(H,A)\cap (G,B)=(G,B)\cap (H,A)$
    \item $(H,A)\cup \Phi=(H,A)$ \label{prop1-5}
    \item $(H,A)\cap \Phi=\Phi$ \label{prop1-6}
    \item $[(H,A)^c]^c=(H,A)$
\end{enumerate}
\end{prop}
%-----------------------------------------------------------------------------------
For any two $NSSs$ $(H,A)$, $(G,B)$  and $(K,C) $over the same
universe $X$, we have the following propositions.
%----------------------------------------------------------------------------------
\begin{prop}\cite{maji-2013}
\begin{enumerate}[(1)]
    \item $(H,A)\cup [(G,B)\cup (K,C)]=[(H,A)\cup (G,B)]\cup (K,C).$
    \item $(H,A)\cap [(G,B)\cap (K,C)]=[(H,A)\cap (G,B)]\cap (K,C).$
    \item $(H,A)\cup [(G,B)\cap (K,C)]=[(H,A)\cup (G,B)]\cap [(H,A)\cup (K,C)].$
    \item $(H,A)\cap [(G,B)\cup (K,C)]=[(H,A)\cap (G,B)]\cup [(H,A)\cap (K,C)].$
\end{enumerate}
\end{prop}
%-----------------------------------------------------------------------------------
\begin{defn}\cite{maji-2013}\label{ns-and} Let $(H,A)$ and $(G,B)$ be two $NSSs$ over
the common universe $X$. Then  'AND' operation on them is denoted by
'$(H,A)\bigwedge(G,B)$'  and is defined by $(H,A)\bigwedge
(G,B)=(K,A\times B)$, where  the truth-membership,
indeterminacy-membership and falsity-membership of $(K,A\times B)$
are as follow.
\begin{eqnarray}
T_{K(\alpha,\beta)}(m)&=&min(T_{H(e)}(m),T_{G(e)}(m))\nonumber\\
I_{K(\alpha,\beta))}(m)&=&\frac{I_{H(e)}(m)+I_{G(e)}(m)}{2} \nonumber\\
F_{K(\alpha,\beta))}(m)&=&max(F_{H(e)}(m),F_{G(e)}(m)), \forall
\alpha\in A, \forall b\in B\nonumber
\end{eqnarray}
\end{defn}
%----------------------------------------------------------------------------------
\begin{defn}\cite{maji-2013}\label{ns-or} Let $(H,A)$ and $(G,B)$ be two $NSSs$ over
the common universe $X$. Then  'OR' operation on them is denoted by
'$(H,A)\bigvee(G,B)$'  and is defined by $(H,A)\bigvee(G,B)=(O,A\times B)$, where  the truth-membership,
indeterminacy-membership and falsity-membership of $(O,A\times B)$
are as follow.
\begin{eqnarray*}
T_{O(\alpha,\beta))}(m)&=&max(T_{H(e)}(m),T_{G(e)}(m)),\\
I_{O(\alpha,\beta))}(m)&=&\frac{I_{H(e)}(m)+I_{G(e)}(m)}{2},\\
F_{O(\alpha,\beta))}(m)&=&min(F_{H(e)}(m),F_{G(e)}(m)), \forall  \alpha\in A, \forall b\in B\nonumber\\
\end{eqnarray*}

\end{defn}
%----------------------------------------------------------------------------------
\subsection*{Notes on neutrosophic soft sets \cite{maji-2013}}
In this section,  we  verify that some propositions in the  study of
Maji \cite{maji-2013} are incorrect by counterexamples.

\begin{enumerate}
  \item If Definition  \eqref{ns-subset} is true, then Definition \eqref{ns-empty} is incorrect.
  \item   Proposition \eqref{prop1}-\eqref{prop1-5} and \eqref{prop1-6}, $(F,A)\cap \Phi =
\Phi$ and  $(F,A)\cup \Phi = (F,A)$  are incorrect.
\end{enumerate}

We verify these notes by  counterexamples.

\begin{exmp}Let us consider  neutrosophic soft set $(F,A)$ in
Example \eqref{examplemaji} and null neutrosophic soft set $\Phi$.
If Definition \eqref{ns-subset} is true, it is required that null
soft set is  neutrosophic soft subset of all  neutrosophic soft
sets. But, since $ T_{\Phi(beautiful)}(h_1)\leq
T_{F(beautiful)}(h_1) $ and $I_{\Phi(beautiful)}(h_1)\leq
I_{F(beautiful)}(h_1)$ but $F_{\Phi(beautiful)}(h_1)\not\geq
F_{F(beautiful)}(h_1)$, $\Phi\not\subseteq (F,A)$.
\end{exmp}
%------------------------------------------------------------------
\begin{exmp}Let us consider  neutrosophic soft set $(F,A)$ in
Example \eqref{examplemaji} and null neutrosophic soft set $\Phi$.
Then,
\begin{eqnarray}
(F,A)\cap \Phi&=&\{e_1 =\{\langle h_1, 0, 0.3, 0.3 \rangle,\langle
h_2, 0, 0.35, 0.6 \rangle,\nonumber\\
&& \langle h_3, 0, 0.1,0.3 \rangle,\langle h_4, 0, 0.15, 0.2
\rangle,\langle h_5, 0, 0.1, 0.3 \rangle\},\nonumber\\
&&e_2=\{\langle h_1, 0, 0.15, 0.5 \rangle,\langle h_2, 0, 0.2, 0.3
\rangle,\langle h_3, 0, 0.05, 0.2 \rangle, \nonumber\\
&&\langle h_4, 0, 0.05, 0.3 \rangle, \langle h_5, 0, 0.15, 0.6
\rangle\},\nonumber\\
&&e_3=\{\langle h_1,0, 0.2, 0.3 \rangle ,\langle h_2, 0, 0.35, 0.2
\rangle,\langle h_3, 0, 0.1, 0.5 \rangle,\nonumber \\
&&\langle h_4, 0, 0.1, 0.6 \rangle,\langle h_5, 0, 0.15, 0.4
\rangle\}, \nonumber\\
&&e_5=\{\langle h_1, 0, 0.3, 0.4 \rangle,\langle h_2, 0, 0.45, 0.6
\rangle,\langle h_3, 0, 0.3, 0.4 \rangle,\nonumber\\
&&\langle h_4, 0, 0.4, 0.6 \rangle, \langle h_5, 0, 0.25, 0.7
\rangle\}\}.\nonumber\\
&\not =&\Phi\nonumber
\end{eqnarray}
and
\begin{eqnarray}
(F,A)\cup \Phi&=&\{e_1 =\{\langle h_1, 0.5, 0.3, 0 \rangle,\langle
h_2, 0.4 0.35, 0 \rangle,\nonumber\\
&& \langle h_3, 0.6, 0.1,0 \rangle,\langle h_4, 0.7, 0.15, 0
\rangle,\langle h_5, 0.8, 0.1, 0 \rangle\},\nonumber\\
&&e_2=\{\langle h_1, 0.6, 0.15, 0 \rangle,\langle h_2, 0.7, 0.2, 0
\rangle,\langle h_3, 0.8, 0.05, 0 \rangle, \nonumber\\
&&\langle h_4, 0.7, 0.05, 0 \rangle, \langle h_5, 0.8, 0.15, 0
\rangle\},\nonumber\\
&&e_3=\{\langle h_1,0.7, 0.2, 0 \rangle ,\langle h_2, 0.6, 0.35, 0
\rangle,\langle h_3, 0.7, 0.1, 0\rangle,\nonumber \\
&&\langle h_4, 0.5, 0.1, 0 \rangle,\langle h_5, 0.7, 0.15, 0
\rangle\}, \nonumber\\
&&e_5=\{\langle h_1, 0.8, 0.3, 0 \rangle,\langle h_2, 0.7, 0.45, 0
\rangle,\langle h_3, 0.7, 0.3, 0 \rangle,\nonumber\\
&&\langle h_4, 0.7, 0.4, 0 \rangle, \langle h_5, 0.9, 0.25, 0
\rangle\}\}.\nonumber\\
&\not =&(F,A)\nonumber
\end{eqnarray}
\end{exmp}
%***********************************************************************************
%% section. Neutrosophic Soft Sets
\section{Neutrosophic soft sets}
In this section, we will redefine the neutrosophic soft
set based on paper of \c{C}a\u{g}man \cite{cagman-2014}.
\begin{defn}\label{neutrosophicsoftset}
A neutrosophic soft set (or namely \emph{ns}-set) $f$ over $X$ is a neutrosophic set
valued function from $E$ to $N(X)$.  It can be written as
$$
f=\Big\{\big(e,\{\langle x,T_{f(e)}(x),I_{f(e)}(x),
F_{f(e)}(x)\rangle:x\in X\}\big): e\in E\Big\}
$$
where, $N(X)$ denotes all neutrosophic sets over $X$. Note that if
$f(e)=\big\{\langle x,0,1,1\rangle: x\in X\big\}$, the element $(e,
f(e))$ is not appeared in the neutrosophic soft set $f$.Set of all
\emph{ns}-sets over $X$ is denoted by $\mathbb{NS}$.
\end{defn}
%-----------------------------------------------------------------------------------
\begin{defn}\label{neutrosophicssubset}Let $f,g\in \mathbb{NS}$. $f$ is said to
be neutrosophic soft subset of $g$, if
$T_{f(e)}(x)\leq T_{g(e)}(x)$, $ I_{f(e)}(x)\geq I_{g(e)}(x)$
$F_{f(e)}(x) \geq F_{g(e)}(x)$, $\forall e\in E$, $\forall x\in U$.
We denote it by $f \sqsubseteq g$. $f$ is said to be
neutrosophic soft super set of $g$ if $g$ is a neutrosophic
soft subset of $f$. We denote it by $f\sqsupseteq g$.

If $f$ is neutrosophic
soft subset of $g$ and $g$ is neutrosophic soft subset of
$f$. We denote it $f=g$
\end{defn}
%-----------------------------------------------------------------
\begin{defn}\label{ns-empty} Let $f\in \mathbb{NS}$. If $T_{f(e)}(x)=0$
and $I_{f(e)}(x)=F_{f(e)}(x)=1$ for all $e\in E$
and for all $x\in X$, then $f$ is called  null \emph{ns}-set and
denoted by $\tilde\Phi$.
\end{defn}
%-----------------------------------------------------------------
\begin{defn}\label{universal} Let $f\in \mathbb{NS}$.If $T_{f(e)}(x)=1$ and
$I_{f(e)}(x)=F_{f(e)}(x)=0$ for all $e\in E$
and for all $x\in X$, then $f$ is called  universal \emph{ns}-set
and denoted by $\tilde X$.
\end{defn}
%---------------------------------------------------------------------
\begin{defn}\label{uni-int}Let $f,g\in \mathbb{NS}$. Then union and intersection of
ns-sets $f$ and $g$ denoted by $f\sqcup g$ and $f\sqcap g$ respectively,  are defined by as follow
\begin{eqnarray*}
f\sqcup g & = & \Big\{\big(e,\{\langle x,T_{f(e)}(x)\vee
T_{g(e)}(x),
I_{f(e)}(x)\wedge I_{g(e)}(x),\\
&&F_{f(e)}(x)\wedge F_{g(e)}(x)\rangle:x\in X\}\big):e\in E\Big\}.
\end{eqnarray*}
and
\emph{ns}-intersection of $f$ and $g$ is defined as
\begin{eqnarray*}
f\sqcap g & = & \Big\{\big(e,\{\langle x,T_{f(e)}(x)\wedge
T_{g(e)}(x),
I_{f(e)}(x)\vee I_{g(e)}(x),\\
&&F_{f(e)}(x)\vee F_{g(e)}(x)\rangle:x\in X\}\big):e\in E\Big\}.
\end{eqnarray*}
\end{defn}
%-------------------------------------------------------------------
\begin{defn}\label{ns-complement} Let $f,g\in \mathbb{NS}$. Then complement of ns-set $f$,
 denoted by $f^{\tilde c}$, is defined as follow
$$
f^{\tilde c}=\Big\{\big(e,\{\langle
x,F_{f(e)}(x),1-I_{f(e)}(x),T_{f(e)}(x)\rangle :x\in X\}\big):e\in
E\Big\}.
$$
\end{defn}
%----------------------------------------------------------------------------------
\begin{prop}Let $f,g,h\in\mathbb{NS}$. Then,
\begin{enumerate}[\it i.]
\item $\tilde \Phi\sqsubseteq f$
\item $f\sqsubseteq \tilde X$
\item $f\sqsubseteq f $
\item $f\sqsubseteq g $ and $g\sqsubseteq h \Rightarrow$ $f\sqsubseteq h$
\end{enumerate}
\end{prop}
\proof The proof is obvious from Definition
\eqref{neutrosophicssubset}, \eqref{ns-empty} and Definition
\eqref{universal}.
%-----------------------------------------------------------------------------------
\begin{prop}Let $f\in\mathbb{NS}$. Then
\begin{enumerate}[\it i.]
\item $\tilde \Phi^{\tilde c}=\tilde X$
\item $\tilde X^{\tilde c}=\tilde\Phi $
\item $(f^{\tilde c})^{\tilde c}=f$.
\end{enumerate}
\end{prop}
\proof The proof is clear from Definition \eqref{ns-empty},
\eqref{universal} and \eqref{ns-complement}.
%-----------------------------------------------------------------------------------
\begin{thm}
Let $f,g,h\in\mathbb{NS}$. Then,
\begin{enumerate}[\it i.]
\item $f\sqcap f=f$ and $f\sqcup f=f$
\item $f\sqcap g=g\sqcap f$ and $f\sqcup g=g\sqcup f$
\item $f\sqcap\tilde\Phi=\tilde\Phi$ and $f\sqcap\tilde X=f$
\item $f\sqcup\tilde\Phi=f$ and $f\sqcup\tilde X=\tilde X$
\item $f\sqcap(g\sqcap h)=(f\sqcap g)\sqcap h$ and $f\sqcup(g\sqcup h)=(f\sqcup g)\sqcup h$
\item $f\sqcap(g\sqcup h)=(f\sqcap g)\sqcup(f\sqcap h)$ and
$f\sqcup(g\sqcap h)=(f\sqcup g)\sqcap(f\sqcup h).$
\end{enumerate}
\end{thm}
\proof The proof is clear from definition and operations of
neutrosophic soft sets.
%\proof
%\begin{rem} Note that, $f\sqcap\hat\Phi\not=\hat\Phi$, $f\sqcap\tilde
%X\not=f$, $f\sqcup\hat\Phi\not=f$ and $f\sqcup\tilde X\not=\tilde
%X$.
%\end{rem}
%-----------------------------------------------------------------------------------
\begin{thm}\label{t-ns-demorgan}
Let $f,g\in\mathbb{NS}$. Then, De Morgan's law is valid.
\begin{enumerate}[\it i.]
\item $(f\sqcup g)^{\tilde c}=f^{\tilde c}\sqcap g^{\tilde c}$
\item $(f\sqcup g)^{\tilde c}=f^{\tilde c}\sqcap g^{\tilde c}$
\end{enumerate}
\end{thm}
\proof $f,g\in\mathbb{NS}$ is given.
\begin{enumerate}[\it i.]
\item From Definition \ref{ns-complement}, we have
\begin{eqnarray*}
(f\sqcup g)^{\tilde c} & = & \Big\{\big(e,\{\langle
x,T_{f(e)}(x)\vee T_{g(e)}(x),
I_{f(e)}(x)\wedge I_{f(e)}(x),\\
&&F_{f(e)}(x)\wedge F_{f(e)}(x)\rangle:x\in X\}\big):e\in E\Big\}^{\tilde c}\\
& = & \Big\{\big(e,\{\langle x,F_{f(e)}(x)\wedge F_{f(e)}(x),
1-(I_{f(e)}(x)\wedge I_{f(e)}(x)),\\
&&T_{f(e)}(x)\vee T_{g(e)}(x)\rangle:x\in X\}\big):e\in E \Big\}\\
&&\langle x,F_{f(e)}(x), 1-I_{f(e)}(x),T_{f(e)}(x)\rangle:e\in E\Big\}\\
& = & \Big\{\big(e,\{X\}\big):e\in E\Big\}\\
&\sqcap&\Big\{\big(e,\{\langle x,F_{g(e)}(x), 1-I_{g(e)}(x),T_{g(e)}(x)\rangle:x\in X\}\big):e\in E\Big\}\\
 & = & f^{\tilde c}\sqcap g^{\tilde c}.
\end{eqnarray*}
\item It can be proved similar way (\emph{i.})
\end{enumerate}
%--------------------------------------------------------------
\begin{defn}Let $f,g\in\mathbb{NS}$. Then, difference of $f$ and $g$, denoted by $f\setminus
g$ is defined by the set of ordered pairs
$$
f\setminus g=\Big\{(e,\{\langle x,T_{f\setminus g(e)}(x),
I_{f\setminus g(e)}(x),F_{f \setminus g(e)}(x)\rangle: x\in X\}):
e\in E\Big\}
$$
here, $T_{f\setminus g(e)}(x)$, $I_{f\setminus g(e)}(x)$ and
$F_{f\setminus g(e)}(x)$ are defined by
$$T_{f\setminus g(e)}(x)=\left\{
  \begin{array}{ll}
 T_{f(e)}(x)-T_{g(e)}(x), & T_{f(e)}(x)>T_{g(e)}(x) \\
    0, & otherwise
  \end{array}
\right.
$$
$$I_{f\setminus g(e)}(x)=\left\{
  \begin{array}{ll}
 I_{g(e)}(x)-I_{f(e)}(x), & I_{f(e)}(x)<I_{g(e)}(x) \\
    0, & otherwise
  \end{array}
\right.
$$
$$F_{f\setminus g(e)}(x)=\left\{
  \begin{array}{ll}
 F_{g(e)}(x)-F_{f(e)}(x), & G_{f(e)}(x)<G_{g(e)}(x) \\
    0, & otherwise
  \end{array}
\right.
$$
\end{defn}
%---------------------------------------------------------------------
\begin{defn}\label{or} Let $f,g\in \mathbb{NS}$. Then  'OR' product  of
ns-sets $f$ and $g$ denoted by $f\wedge g$, is defined  as follow
\begin{eqnarray*}
f\bigvee g & = & \Big\{\big((e,e'),\{\langle x,T_{f(e)}(x)\vee
T_{g(e)}(x),
I_{f(e)}(x)\wedge I_{g(e)}(x),\\
&&F_{f(e)}(x)\wedge F_{g(e)}(x)\rangle:x\in X\}\big):(e,e')\in
E\times E\Big\}.
\end{eqnarray*}
\end{defn}
%---------------------------------------------------------------------
\begin{defn}\label{and} Let $f,g\in \mathbb{NS}$. Then  'AND' product  of
ns-sets $f$ and $g$ denoted by $f\vee g$, is defined  as follow
\begin{eqnarray*}
f\bigwedge g & = & \Big\{\big((e,e'),\{\langle x,T_{f(e)}(x)\wedge
T_{g(e)}(x),
I_{f(e)}(x)\vee I_{g(e)}(x),\\
&&F_{f(e)}(x)\vee F_{g(e)}(x)\rangle:x\in X\}\big):(e,e')\in E\times
E\Big\}.
\end{eqnarray*}
\end{defn}
%---------------------------------------------------------------------------
\begin{prop} Let $f,g\in \mathbb{NS}$. Then,
\begin{enumerate}
\item $(f\bigvee g)^{\tilde c}=f^{\tilde c}\bigwedge g^{\tilde c}$
\item $(f\bigwedge g)^{\tilde c}=f^{\tilde c}\bigvee g^{\tilde c}$
\end{enumerate}
\end{prop}
\proof The proof is clear from Definition \eqref{or} and
\eqref{and}.

%***********************************************************************************
%% section. Neutrosophic Soft Sets
\section{Decision making method}
In this section we will construct a decision making  method over the
neutrosophic soft set. Firstly, we will define  some notions that necessary
 to construct algorithm of decision making method.

%A neutrosophic soft set is a mapping from
%parameter set $E$ to set of all  neutrosohic sets over $X$.
%Neutrosophic soft set is the most general situation of fuzzy soft
%sets and intuitionistic fuzzy soft sets. In a fuzzy soft set or
%intuitionistic fuzzy soft set, we always suppose that parameters
%have equal importance. Whereas, some parameters can be much more
%important for us than others. Therefore, firstly we will construct a
%matrix expressing the relative values of a set of parameters. For
%example, what is the relative importance to the management of this
%firm of the cost of equipment as opposed to its ease of operation?
%They are asked to choose whether cost is very much more important,
%rather more important, and as important, and so on down to very much
%less The Analytic Hierarchy Process (AHP) important, than
%operability. Each of these judgments is assigned a number on a
%scale. One common scale (adapted from Saaty) is the one shown in
%Table 1.
%
%Secondly, for neutrosophic set that corresponding to each of
%parameters, we will present a method converting fuzzy set to
%neutrosophic set. Then, we will construct a matrix expressing
%according to the membership degrees of the objects belonging to $X$.
%Next, we will present some basic definitions that will be used for
%decision making method and then we will present decision algorithm.

\begin{defn}\label{rp-matris} Let $X=\{x_1,x_2,...x_m\}$ be an initial
universe, $E=\{e_1,e_2,...e_n\}$ be  a parameter set and $f$ be a
neutrosophic soft set over $X$. Then, according to the Table of
"Saaty Rating Scale" relative parameter matrix $d_E$ is defined as
follow
$$
d_E=\left[
  \begin{array}{cccc}
    1& d_E(e_1,e_2)& \ldots & d_E(e_1,e_n) \\
    d_E(e_2,e_1)& 1& \ldots & d_E(e_2,e_n) \\
    \vdots & \vdots& \vdots & \vdots  \\
    d_E(e_n,e_1)& d_E(e_n,e_2)& \ldots & 1 \\
  \end{array}
\right]
$$
If $d_E(e_i,e_j)=d_{12}$, we can write matrix
$$
d_E=\left[
  \begin{array}{cccc}
    1& d_{11}& \ldots & d_{1n} \\
    d_{21}& 1& \ldots & d_{2n} \\
    \vdots & \vdots& \vdots & \vdots  \\
    d_{n1}& d_{n2}& \ldots & 1 \\
  \end{array}
\right]
$$
Here, $d_{12}$ means that how much important $e_1$ by $e_2$.
 For example, if $e_1$ is much more important by $e_2$, then we can write $d_{12}=5$ from Table 1.

\begin{scriptsize}
$$
\begin{tabular}{|l|l|l|}
\hline Intensity importance & Definition & Explanation \\
\hline
\(1\)& Equal importance & Two factors contribute \\
& &  equally to the objective  \\
\hline \(3\)& Somewhat more important& Experience and judgement\\
& &slightly favour one over
the other \\
\hline \(5\)& Much more important & Experience and judgement\\
& &strongly favour one over
the other  \\
\hline \(7\)& Very much more important& Experience and judgement\\
& &very strongly favour one over the other. \\
& & Its importance is demonstrated in
practice \\
\hline \(9\)& Absolutely more important & The evidence favouring one
over the other\\
& & is of the
highest possible validity.  \\
 \hline
\(2,4,6,8\)& Intermediate
values & When compromise is needed \\
\hline
\end{tabular}
$$
\end{scriptsize}
\begin{center}\label{satty}
\footnotesize{\emph{\emph{Table 1}. The Saaty Rating Scale}}
\end{center}
%$$
%\begin{tabular}{c|cccc}
%  \hline
% \(D_E\) &  \(e_1\) &
%\(e_2\) &...&  \(e_n\) \\
%    \hline \(e_1\)  &
%\(1\)& \(D_E(e_1,e_2)\)& \ldots & \(D_E(e_1,e_n)\) \\
%    \(e_2\) &
%\(D_E(e_2,e_1)\)& \(1\)& \ldots & \(D_E(e_2,e_n)\) \\
%    \(\vdots\)  &
%\(\vdots\)& \(\vdots\)& \(\vdots\) & \(\vdots\) \\
%\(e_n\)  &
%\(D_E(e_n,e_1)\)& \(D_E(e_n,e_2)\)& \ldots & \(1\) \\
%   \hline
%    \end{tabular}
%   $$
%\begin{center}
%\footnotesize{\emph{\emph{Table 1}. The tabular representation of
%the relative parameter matrix $D_E$}}
%\end{center}
\end{defn}
%-------------------------------------------------------------
\begin{defn}Let $f$ be a neutrosophic soft set and $d_E$ be a relative parameter matrix of $f$. Then,
 score of parameter  $e_i$, denoted by $c_i$ and is  calculated as follows
$$c_i=\sum_{j=1}^nd_{ij}$$
\end{defn}
%---------------------------------------------------------------
\begin{defn} Normalized relative parameter matrix ($nd_E$ for short) of relative parameter
matrix $d_E$, denoted by $\hat d$, is defined as follow,
$$
nd_E=\left[
  \begin{array}{cccc}
   \frac{1}{c_1} & \frac{d_{12}}{c_1} & \ldots & \frac{d_{1n}}{c_1}  \\
   \frac{d_{21}}{c_2} & \frac{1}{c_2}& \ldots & \frac{d_{2n}}{c_2} \\
   \vdots & \vdots  &  \ddots & \vdots  \\
   \frac{d_{n1}}{c_n} & \frac{d_{n2}}{c_n} & \ldots & \frac{1}{c_n}
  \end{array}
\right]
$$
if  $\frac{d_{ij}}{c_i}=\hat d_{ij}$, we can write matrix $nd_E$

$$
\hat d=\left[
  \begin{array}{cccc}
   \hat d_{11} & \hat d_{12} & \ldots & \hat d_{1n}  \\
   \hat d_{21}  & \hat d_{22} & \ldots & \hat d_{2n} \\
   \vdots & \vdots  &  \ddots & \vdots  \\
   \hat d_{n1} & \hat d_{n2} & \ldots & \hat d_{nn}
  \end{array}
\right]
$$

\end{defn}

%-------------------------------------------------------------
\begin{defn} Let $f$ be a neutrosophic soft set and $\hat d$ be a normalized
 parameter matrix of $f$. Then, weight of parameter $e_j\in E$, denoted by $w(e_j)$ and is
formulated as follows.
$$
w(e_j)=\frac{1}{|E|}\sum_{i=1}^n\hat d_{ij}
$$
\end{defn}
Now, we construct compare matrices of elements of $X$ in
neutrosophic sets $f(e)$, $\forall e\in E$.
%---------------------------------------------------------------
\begin{defn} Let $E$ be a parameter set and  $f$ be a neutrosophic soft set over $X$. Then, for
all $e\in E$, compression matrices of $f$, denoted
$X_{f(e)}$ is defined  as follow
$$
X_{f(e)}=\left[
         \begin{array}{cccc}
           x_{11} & x_{12}  & \cdots & x_{1m}  \\
           x_{21}  & x_{22}  & \cdots & x_{2m}  \\
           \vdots & \vdots & \ddots & \vdots \\
           x_{m1}  & x_{m2}  & \cdots & x_{mm}  \\
         \end{array}
       \right]
$$

$$
x_{ij}=\frac{\Delta_{T(e)}(x_{ij})+\Delta_{I(e)}(x_{ij})+\Delta_{F(e)}(x_{ij})+1}{2}
$$
such that
\begin{eqnarray*}
\Delta_{T(e)}(x_{ij})&=&T(e)(x_{i})-T(e)(x_{j})\\
\Delta_{I(e)}(x_{ij})&=&I(e)(x_{j})-I(e)(x_{i})\\
\Delta_{F(e)}(x_{ij})&=&F(e)(x_{j})-F(e)(x_{i})\\
\end{eqnarray*}
\end{defn}
%---------------------------------------------------------------
\begin{defn} Let $X_{f(e)}$ be compare matrix for $e\in E$. Then,
weight of $x_j\in X$ related to parameter $e\in E$, denoted by
$W_{f(e)}(x_j)$ is defined  as follow,
$$
W_{f(e)}(x_j)=\frac{1}{|X|}\sum_{i=1}^mx_{ij}
$$
\end{defn}
%--------------------------------------------------------------------
\begin{defn}Let $E$ be a parameter set, $X$ be an initial
universe and $w(e)$ and $W_{f(e)}(x_j)$ be weight of parameter $e$
and membership degree of $x_j$ which related to $e_j\in E$,
respectively. Then, decision set, denoted $D_{E}$, is defined by the
set of ordered pairs
$$
D_E=\{(x_j,F(x_j)): x_j\in X\}
$$
where
$$F(x_j)=\frac{1}{|E|}\sum_{j=1}^nw(e_j)\cdot W_{f(e)}(x_j)$$

Note that, $F$ is a fuzzy set over $X$.
\end{defn}
Now, we construct a neutrosophic soft set  decision making method by
the following algorithm;
%**************************************************************
\section*{Algorithm 1}
%---------------------------------------------------------------
\emph{\textbf{Step 1:}} Input the neutrosophic soft set $f$,\\
\emph{\textbf{Step 2:}} Construct the normalized parameter matrix,\\
\emph{\textbf{Step 3:}} Compute the weight of each  parameters, \\
\emph{\textbf{Step 4:}} Construct the compare matrix for each parameter,\\
\emph{\textbf{Step 5:}} Compute membership degree, for all $x_j\in X$\\
\emph{\textbf{Step 6:}} Construct decision set $D_E$\\
\emph{\textbf{Step 7:}} The optimal decision is to select $x_k=max F(x_j)$.\\

%--------------------------------------------------------------
\begin{exmp} Let $X$ be the set of blouses under consideration and
$E$ is the set of parameters. Each parameters is a neutrosophic word
or sentence involving neutrosophic words. Consider
$E=\{\textrm{bright}, \textrm{cheap}, \textrm{colorful},
\textrm{cotton}\}$. Suppose that, there are five blouses in the
universe $X$ given by $X=\{x_1,x_2,x_3,x_4,x_5\}$. Suppose that,\\

\textbf{Step 1:} Let us consider the decision making problem
involving the neutrosophic soft set in
\cite{bro-13gns}
\footnotesize{
\begin{eqnarray}
F(Bright)&=&\{\langle x_1, .5,.6,.3\rangle,\langle x_2,
.4,.7,.2\rangle,\langle x_3, .6,.2,.3\rangle,\langle x_4,
.7,.3,.2\rangle,\langle x_5, .8,.2,.3\rangle,\}\nonumber\\
F(Cheap)&=&\{\langle x_1, .6,.3,.5\rangle,\langle x_2,
.7,.4,.3\rangle,\langle x_3, .8,.1,.2\rangle,\langle x_4,
.7,.1,.3\rangle,\langle x_5, .8,.3,.4\rangle,\}\nonumber\\
F(Colorful)&=&\{\langle x_1, .7,.4,.3\rangle,\langle x_2,
.6,.1,.2\rangle,\langle x_3, .7,.2,.5\rangle,\langle x_4,
.5,.2,.6\rangle,\langle x_5, .7,.3,.2\rangle,\}\nonumber\\
F(Cotton)&=&\{\langle x_1, .4,.3,.7\rangle,\langle x_2,
.5,.4,.2\rangle,\langle x_3, .7,.4,.3\rangle,\langle x_4,
.2,.4,.5\rangle,\langle x_5, .6,.4,.4\rangle,\}\nonumber
\end{eqnarray}}\\

\normalsize
\textbf{Step 2:}
$$
d_E=\left(
\begin{array}{cccc}
1& 1/3& 5 & 1/3 \\
3 & 1 & 2 & 3 \\
1/5 & 1/2 & 1 & 2 \\
3 & 1/3 & 1/2 & 1 \\
\end{array}
\right)
$$
$c_1=6.67$, $c_2=9$,  $c_3=3.7$ and $c_4=4.88$ and

$$
\hat d_E=\left(
\begin{array}{cccc}
.15& .05 & .75 & .05 \\
.33 & .11 & .22 & .33 \\
.05 & .14 & .27 & .54 \\
.62 & .07 & .10 & .21 \\
\end{array}
\right)
 $$\\

\textbf{Step 3:} From normalized matrix, weight of  parameters are
obtained as $w(e_1)=.29$, $w(e_2)=.09$, $w(e_3)=.34$ and
$w(e_4)=.28$.\\

\textbf{Step 4:} For each parameter, compare matrices
and normalized compare matrices are constructed as follow\\

Let us consider parameter "bright". Then,
\footnotesize
$$X_{f(bright)}=\left[
\begin{array}{ccccc}
.50 & .10 & .25 & .20 & .15 \\
.45 & .50 & .20  & .15  & .10 \\
.75 & .80 & .50 & .45 & .40 \\
.80 & .85 & .55 & .50 & .45 \\
.85 & .90 & .60  & .55 & .50\\
    \end{array}
\right], \quad X_{f(cheap)}=\left[
\begin{array}{ccccc}
.50 & .40 & .15 & .25 & .35 \\
.50 & .50 & .30 & .35 & .45 \\
.85 & .75 & .50 & .60 & .70 \\
.75 & .65 & .40 & .50 & .60 \\
.65 & .55 & .30 & .40 & .50\\
    \end{array}
\right]
$$
and
$$
 X_{f(colorful)}=\left[
\begin{array}{ccccc}
.50 & .35 & .55 & .65 & .40 \\
.65 & .50 & .65 & .18 & .55 \\
.50 & .35 & .50 & .65 & .40 \\
.35 & .30 & .15 & .50 & .25 \\
.40 & .45 & .60 & .75 & .50 \\
    \end{array}
\right], \quad
 X_{f(cotton)}=\left[
\begin{array}{ccccc}
.50 & .25 & .35 & .50 & .15 \\
.75 & .50 & .60 & .75 & .40 \\
.65 & .40 & .50 & .65 & .30 \\
.50 & .25 & .35 & .50 & .15 \\
.85 & .60 & .70 & .85 & .50\\
    \end{array}
\right]
$$

\normalsize
\textbf{Step 5:} For all $x_j\in X$ and $e\in E$,\\

$
W_{f(bright)}(x_1)=.67,\; W_{f(bright)}(x_2)=.63,\; W_{f(bright)}(x_3)=.42,\\
W_{f(bright)}(x_4)=.37, \; W_{f(bright)}(x_5)=.32
$\\

$
W_{f(cheap)}(x_1)=.80, \;W_{f(cheap)}(x_2)=.57, \;W_{f(cheap)}(x_3)=.33,\\
 W_{f(cheap)}(x_4)=.42, \;W_{f(cheap)}(x_5)=.52
$\\

$
W_{f(colorful)}(x_1)=.48,\; W_{f(colorful)}(x_2)=.39, \; W_{f(colorful)}(x_3)=.49, \\
W_{f(colorful)}(x_4)=.55, \; W_{f(colorful)}(x_5)=.42
$\\

$
W_{f(cotton)}(x_1)=.65, \; W_{f(cotton)}(x_2)=.40, \; W_{f(cotton)}(x_3)=.50, \\
 W_{f(cotton)}(x_4)=.65, \; W_{f(cotton)}(x_5)=.30
$
\\

\textbf{Step 6:} By using  step 3 and step 5, $D_E$ is constructed as follow
$$D_E=\{(x_1,0.15),(x_2,0.12),(x_3,0.11),(x_4,0.13),(x_5,0.09)\}$$

\textbf{Step 7:}  Note that, membership degree of $x_1$ is greater than the other. Therefore,
optimal decision is $x_1$ for this decision making problem.
\end{exmp}
%*************************************************************
\section{Group decision making}
%****************************************************************
In this section,  we constructed a group decision making method using intersection of neutrosophic soft sets  and Algorithm 1.\\

%----------------------------------------------------------------
Let $X= \{x_1, x_2,..., x_n \}$ be an initial universe and let $d =
\{d^1, d^2,..., d^m\}$ be a decision maker set and
$E=\{e_1,e_2,...,e_k\}$ be a set of  parameters. Then, this method
can be described by
the following steps \\
\section*{Algorithm 2}
%-----------------------------------------------------------------------------------

\emph{\textbf{Step 1:}} Each decision-maker $d^i$ construct own
neutrosophic soft set, denoted by  $f_{d_i}$, over  $U$ and
parameter set $E$.\\
\emph{\textbf{Step 2:}} Let for $p,r\leq k$,  $[d^i_{pr}]$ a
relative parameter matrix  of decision-maker $d^i\in D$ based on the
Saaty Rating Scale. Decision-maker $d^i$ gives his/her evaluations
separately and independently according to his/her own preference
based on Saaty Rating Scale. In this way, each decision-maker $d^i$
presents a relative parameter matrix.

$$[d^i_{pr}]=\left(
               \begin{array}{cccc}
                 d^i_{11} & d^i_{12} & \cdots & d^i_{1k} \\
                 d^i_{21} & d^i_{22} & \cdots & d^i_{2k} \\
                 \vdots & \vdots & \ddots & \vdots \\
                 d^i_{k1} & d^i_{k2} & \vdots & d^i_{kk} \\
               \end{array}
             \right)
$$
here $d^i_{pr}$ is equal $d_E(e_p, e_r)$ that in Definition
\eqref{rp-matris}.\\
\emph{\textbf{Step 3:}} Arithmetic mean  matrix is constructed by
using the the relative parameter matrix of each decision-maker
$d^i$. It will be denoted by  $[i_{pr}]$ and will be computed as in
follow
$$
i_{pr}=\frac{1}{|d|}\sum_{i=1}^md^i_{pr}
$$\\
\emph{\textbf{Step 4:}} Normalized parameter matrix, is constructed
using the arithmetic mean matrix $[i_{pr}]$, it will be shown $[\hat
i_{pr}]$ and weight of each parameter $e_i\in E$ ($w(e_i)$) is
computed.\\
\emph{\textbf{Step 5:}} Intersection of neutrosophic soft sets (it
will be denoted by $I_{f_{d_i}}$)
 which are constructed by decision makers is found.
$$
I_{f_{d}}=\bigcap_{i=1}^mf_{d^i}
$$\\
\emph{\textbf{Step 6:}} Based on the matrix $I_{f_{d}}$,  for each
element of $e\in E$ compare matrix, denoted by $I_{f_{d}(e)}$ is
constructed.\\
\emph{\textbf{Step 7:}} By the  $I_{f_{d}(e)}$, weight of  each
element of $X$ denoted by  $W_{I_{f_{d}(e)}}(x_i)$, are  computed.\\
\emph{\textbf{Step 8:}}  Decision set $D_E$ is constructed by using
values of $w(e)$ and $W_{I_{f_{d^i}}}(x)$. Namely;

$$
D_E=\{(x_i,F(x_i)): x_i\in X\}
$$
and
$$
F(x_i)=\frac{1}{|E|}\sum_{j=1}^nw(e_j)\cdot W_{I_f(e)}(x_i)
$$\\
\emph{\textbf{Step 9:}} From the decision set,  $x_k=max F(x_i)$ is
selected as optimal decision.
\begin{exmp}

\pagebreak

\end{exmp} Assume that a company wants to fill a position.
There are 6 candidates who fill in a form in order to apply formally
for the position. There are three decision makers; one of them is
from the department of human resources and the others is from the
board of directors. They want to interview the candidates, but it is
very difficult to make it all of them. Let $d=\{d_1,d_2,d_3\}$ be a
decision makers set, $X=\{x_1,x_2,x_3,x_4,x_5\}$ be set of
candidates and
 $E =\{e_1,e_2,e_3,e_4\}$ be a parameter
set such that  parameters $e_1,e_2,e_3$ and $e_4$  stand for
 ''experience'',''computer knowledge",  ''higher education'' and ''good health"'', respectively.\\

\emph{\textbf{Step:1}} Let each decision maker construct
neutrosophic soft sets over $X$ by own interview:\\
\footnotesize
 $ f_{d^1}=\left\{
\begin{array}{c}
f_{d^1}(e_1)=\{\langle x_1, .4,.2,.7\rangle,\langle x_2,
.5,.6,.2\rangle,\langle x_3, .7,.3,.3\rangle,\langle x_4,
.6,.5,.4\rangle,\langle x_5, .3,.5,.5\rangle\},\\
f_{d^1}(e_2)=\{\langle x_1, .3,.5,.2\rangle,\langle x_2,
.4,.4,.3\rangle,\langle x_3, .5,.7,.8\rangle,\langle x_4,
.7,.1,.3\rangle,\langle x_5, .6,.3,.2\rangle\},\\
f_{d^1}(e_3)=\{\langle x_1, .7,.4,.3\rangle,\langle x_2,
.6,.1,.5\rangle,\langle x_3, .5,.2,.4\rangle,\langle x_4,
.2,.2,.6\rangle,\langle x_5, .3,.3,.6\rangle\},\\
f_{d^1}(e_4)=\{\langle x_1, .7,.3,.5\rangle,\langle x_2,
.3,.5,.3\rangle,\langle x_3, .2,.4,.3\rangle,\langle x_4,
.4,.2,.5\rangle,\langle x_5, .5,.2,.6\rangle\}
\end{array}\right\}
$
\\
\\
\\
$f_{d^2}=\left\{
\begin{array}{c}
f_{d^2}(e_1)=\{\langle x_1, .5,.2,.3\rangle,\langle x_2,
.3,.5,.6\rangle,\langle x_3, .4,.3,.3\rangle,\langle x_4,
.2,.5,.4\rangle,\langle x_5, .5,.5,.5\rangle\},\\
f_{d^2}(e_2)=\{\langle x_1, .5,.4,.6\rangle,\langle x_2,
.7,.2,.5\rangle,\langle x_3, .6,.3,.5\rangle,\langle x_4,
.7,.2,.3\rangle,\langle x_5, .6,.4,.2\rangle\},\\
f_{d^2}(e_3)=\{\langle x_1, .6,.2,.5\rangle,\langle x_2,
.4,.4,.6\rangle,\langle x_3, .2,.5,.4\rangle,\langle x_4,
.3,.5,.4\rangle,\langle x_5, .3,.3,.6\rangle\},\\
f_{d^2}(e_4)=\{\langle x_1, .3,.4,.5\rangle,\langle x_2,
.4,.3,.2\rangle,\langle x_3, .4,.4,.3\rangle,\langle x_4,
.4,.2,.5\rangle,\langle x_5, .2,.5,.6\rangle\}
\end{array}\right\}
$ and\\
$ f_{d^3}=\left\{
\begin{array}{c}
f_{d^3}(e_1)=\{\langle x_1, .4,.5,.7\rangle,
        \langle x_2,.5,.3,.4\rangle,
        \langle x_3, .7,.3,.5\rangle,
        \langle x_4,.4,.5,.3\rangle,
        \langle x_5, .7,.8,.6\rangle\},\\
f_{d^3}(e_2)=\{\langle x_1, .6,.2,.6\rangle,
        \langle x_2,.4,.3,.5\rangle,
        \langle x_3, .5,.4,.7\rangle,
        \langle x_4,.3,.1,.5\rangle,
        \langle x_5, .4,.3,.1\rangle\},\\
f_{d^3}(e_3)=\{\langle x_1, .4,.3,.2\rangle,
        \langle x_2,.6,.7,.2\rangle,
        \langle x_3, .3,.5,.2\rangle,
        \langle x_4,.6,.6,.4\rangle,
        \langle x_5, .6,.5,.5\rangle\},\\
f_{d^3}(e_4)=\{\langle x_1, .5,.3,.1\rangle,
        \langle x_2,.2,.5,.2\rangle,
        \langle x_3, .5,.5,.4\rangle,
        \langle x_4,.5,.2,.5\rangle,
        \langle x_5, .5,.3,.6\rangle\}
\end{array}\right \}$
\\
 \normalsize

\emph{ \textbf{Step 2:}} Relative parameter matrix of each decision
maker are as in follow
$$[d^1_{pr}]=\left[
               \begin{array}{cccc}
                 1 & 3 & 1/5 & 2 \\
                 1/3 & 1 & 3 & 6 \\
                 5 & 1/3 & 1 & 1/5 \\
                 1/2 & 1/6 & 5 & 1 \\
               \end{array}
             \right]
\quad
[d^2_{pr}]=\left[
               \begin{array}{cccc}
                 1 & 5 & 1/7 & 2 \\
                 1/5 & 1 & 1/2 & 6 \\
                 7 & 2 & 1 & 1/3 \\
                 1/2 & 1/6 & 3 & 1 \\
               \end{array}
             \right]
$$
and
$$
[d^3_{pr}]=\left[
               \begin{array}{cccc}
                 1 & 3 & 1/3 & 4 \\
                 1/3 & 1 & 1/3 & 1/6 \\
                 3& 3 & 1 & 1/2 \\
                 1/4 & 6 & 2 & 1 \\
               \end{array}
             \right]
$$\\

\emph{\textbf{Step 3:}} $[i_{pr}]$ can be obtained as follow

$$
[i_{pr}]=\left[
               \begin{array}{cccc}
                 1 & 3.67 &.23  &2.67  \\
               .29 & 1 & 1.28  &4.06  \\
                 5& 1.78 & 1 &.34  \\
                  .42& 4.06 &3.33  & 1 \\
               \end{array}
             \right]
$$\\

\emph{\textbf{Step 4:}} $[\hat i_{pr}]$ and weight of each parameter
can be obtained as follow

$$[\hat i_{pr}]=\left[
               \begin{array}{cccc}
                 .13& .49 &.03  &.35  \\
                  .04& .15 & .19  & .61  \\
                 .62& .22 & .12 &.04  \\
                  .05& .46 &.38  & .11 \\
               \end{array}
             \right]
$$
and $w(e_1)=.21$,  $w(e_2)=0.33$,  $w(e_3)=.18$  $w(e_4)=.28$.\\

\emph{\textbf{Step 5:}} Intersection of neutrosophic soft sets
$f_{d^1},f_{d^2}$ and $f_{d^3}$ is as follow;
\\
\\
\footnotesize $  I_{f_d}=\left\{
\begin{array}{c}
I_{f_d}(e_1)=\{\langle x_1, .4,.5,.7\rangle,
        \langle x_2,.3,.6,.6\rangle,
        \langle x_3, .4,.3,.5\rangle,
        \langle x_4,.2,.5,.5\rangle,
        \langle x_5, .3,.8,.6\rangle\},\\
I_{f_d}(e_2)=\{\langle x_1, .3,.5,.6\rangle,
        \langle x_2,.4,.4,.5\rangle,
        \langle x_3, .5,.7,.8\rangle,
        \langle x_4,.3,.2,.5\rangle,
        \langle x_5, .4,.4,.2\rangle\},\\
I_{f_d}(e_3)=\{\langle x_1, .6,.5,.5\rangle,
        \langle x_2,.4,.7,.6\rangle,
        \langle x_3, .2,.5,.4\rangle,
        \langle x_4,.2,.6,.6\rangle,
        \langle x_5, .3,.5,.6\rangle\},\\
I_{f_d}(e_4)=\{\langle x_1, .3,.4,.5\rangle,
        \langle x_2,.2,.5,.3\rangle,
        \langle x_3, .2,.5,.4\rangle,
        \langle x_4,.4,.2,.5\rangle,
        \langle x_5, .2,.5,.6\rangle\}
\end{array}\right \}$
\\
\normalsize

\emph{\textbf{Step 6:}} For each parameter, compare matrices of
elements of $X$ are obtained as in follow;
$$
I_{f_d(e_1)}=\left[
               \begin{array}{ccccc}
                 .50 & .55 & .30 & .45 &.65 \\
                 .45 & .50 & .25 & .40 &.60 \\
                 .70& .75 & .50 & .35 & .85\\
                 .55 & .60 & .65 & .50 &.70\\
                 .65 & .40 & .15 & .30 &.70\\
               \end{array}
             \right],
\quad I_{f_d(e_2)}=\left[
               \begin{array}{ccccc}
                 .50 & .35 & .60 & .30 &.20 \\
                 .65 & .50 & .75 & .35 &.35 \\
                 .40 & .25 & .50 & .20 &.10\\
                 .70 & .65 & .80 & .50 &.40\\
                 .80 & .65 & .90 & .60 &.50\\
               \end{array}
             \right]
$$
and
$$
I_{f_d(e_3)}=\left[
               \begin{array}{ccccc}
                 .50 & .75 & .65 & .80 &.70 \\
                 .25 & .50 & .40 & .55 &.45 \\
                 .35& .60 & .50 & .65 & .55\\
                 .20 & .45 & .35 & .50 &.45\\
                 .30 & .55 & .45 & .55 &.50\\
               \end{array}
             \right],
\quad I_{f_d(e_4)}=\left[
               \begin{array}{ccccc}
                 .50 & .50 & .55 & .35 &.65 \\
                 .50 & .50 & .45 & .40 &.65 \\
                 .45 & .55 & .50 & .30 &.60\\
                 .65 & .60 & .70 & .50 &.80\\
                 .35 & .35 & .40 & .20 &.50\\
               \end{array}
             \right]
$$\\

\emph{\textbf{Step 7:}} Membership degrees of  elements of $X$
related to each parameter $e\in E$ are obtained as follow;

 $W_{f_{d}(e_1)}(x_1)=.57$,
$W_{f_{d}(e_1)}(x_2)=.56$, $W_{f_{d}(e_1)}(x_3)=.37$,
$W_{f_{d}(e_1)}(x_4)=.40$ and
$W_{f_{d}(e_1)}(x_5)=.66$\\

$W_{f_{d}(e_2)}(x_1)=.61$, $W_{f_{d}(e_2)}(x_2)=.48$,
$W_{f_{d}(e_2)}(x_3)=.71$, $W_{f_{d}(e_2)}(x_4)=.39$ and
$W_{f_{d}(e_2)}(x_5)=.31$\\

$W_{f_{d}(e_3)}(x_1)=.32$, $W_{f_{d}(e_3)}(x_2)=.57$,
$W_{f_{d}(e_3)}(x_3)=.47$, $W_{f_{d}(e_3)}(x_4)=.61$ and
$W_{f_{d}(e_3)}(x_5)=.53$\\

$W_{f_{d}(e_4)}(x_1)=.49$, $W_{f_{d}(e_4)}(x_2)=.50$,
$W_{f_{d}(e_4)}(x_3)=.52$,$W_{f_{d}(e_4)}(x_4)=.35$ and
$W_{f_{d}(e_4)}(x_5)=.64$\\

\emph{\textbf{Step 8:}}
\begin{eqnarray*}
F(x_1)&=&\frac{1}{|E|}\sum_{j=1}^nw(e_j)\cdot W_{{f_{d}(e_j)}}(x_1)\\
&=&\frac{1}{4}(.21\cdot.57+.33\cdot.61+.18\cdot.32+.28\cdot.49)\\
&=&.126
\end{eqnarray*}
similarly  $F(x_2)=.130$, $F(x_3)=.136$, $F(x_4)=.105$ and
$F(x_5)=.129$. Then, we get
$$
D_E=\{(x_1,.126),(x_1,.130),(x_1,.136),(x_1,.105),(x_1,.129)\}
$$\\

\emph{\textbf{Step 9:}}  Note that, membership degree of $x_3$ is
greater than membership degrees of the others. Therefore, optimal
decision is $x_3$ for this decision making problem.
%**********************************************************************************
\section{Conclusion}
%***********************************************************************************
In this paper, we firstly investigate  neutrosophic soft sets given
paper of Maji \cite{maji-2013} and then we redefine notion of
neutrosophic soft set and neutrosophic soft set  operations.
Finally, we present two applications of neutrosophic soft sets in
decision making problem.
%% bibliography


\begin{thebibliography}{00}
\bibitem{deli-14ivnss} Ý. Deli,Interval-valued neutrosophic soft sets ant its decision making, arxiv:1402.3130
\end{thebibliography}
\end{document}